\documentclass[10pt,twocolumn,letterpaper]{article}

\usepackage{iccv}
\usepackage{times}
\usepackage{epsfig}
\usepackage{graphicx}
\usepackage{amsmath}
\usepackage{amssymb}

\usepackage{multirow}
\usepackage{xcolor}
\usepackage{array}
\usepackage{subfigure}
\usepackage{boldline}

\usepackage[breaklinks=true,bookmarks=false]{hyperref}

\iccvfinalcopy 

\def\iccvPaperID{1142} 

\def\head#1{\noindent\textbf{#1}}
\newcolumntype{P}[1]{>{\centering\arraybackslash}p{#1}}

\ificcvfinal\fi
\begin{document}

\title{ThunderNet: Towards Real-time Generic Object Detection on Mobile Devices}

\author{Zheng Qin\thanks{Equal contribution.}\hspace{4pt}\thanks{This work was done when Zheng Qin was an intern at Megvii Inc.}\hspace{4pt}$^{1}$\hspace{8pt}Zeming Li\footnotemark[1]\hspace{4pt}$^{2}$\hspace{8pt}Zhaoning Zhang$^{1}$\hspace{8pt}Yiping Bao$^{2}$\hspace{8pt}Gang Yu$^{2}$\thanks{Corresponding author.}\hspace{12pt}Yuxing Peng$^{1}$\hspace{8pt}Jian Sun$^{2}$\\
$^{1}$\hspace{2pt}National University of Defense Technology\hspace{45pt}$^{2}$\hspace{2pt}Megvii Inc. (Face++)
}

\maketitle
\ificcvfinal\thispagestyle{empty}\fi


\begin{abstract}
Real-time generic object detection on mobile platforms is a crucial but challenging computer vision task.
Prior lightweight CNN-based detectors are inclined to use one-stage pipeline.
In this paper, we investigate the effectiveness of two-stage detectors in real-time generic detection and propose a lightweight two-stage detector named ThunderNet.
In the backbone part, we analyze the drawbacks in previous lightweight backbones and present a lightweight backbone designed for object detection.
In the detection part, we exploit an extremely efficient RPN and detection head design.
To generate more discriminative feature representation, we design two efficient architecture blocks, Context Enhancement Module and Spatial Attention Module.
At last, we investigate the balance between the input resolution, the backbone, and the detection head.
Benefit from the highly efficient backbone and detection part design, 
ThunderNet surpasses previous lightweight one-stage detectors with only 40\% of the computational cost on PASCAL VOC and COCO benchmarks.
Without bells and whistles, ThunderNet runs at 24.1 fps on an ARM-based device with 19.2 AP on COCO.
To the best of our knowledge, this is the first real-time detector reported on ARM platforms.
Our code and models are available at \url{https://github.com/qinzheng93/ThunderNet}.
\end{abstract}
  

\section{Introduction}
\label{section:introduction}


\begin{figure}[!t]
\centering
\includegraphics[width=0.47\textwidth]{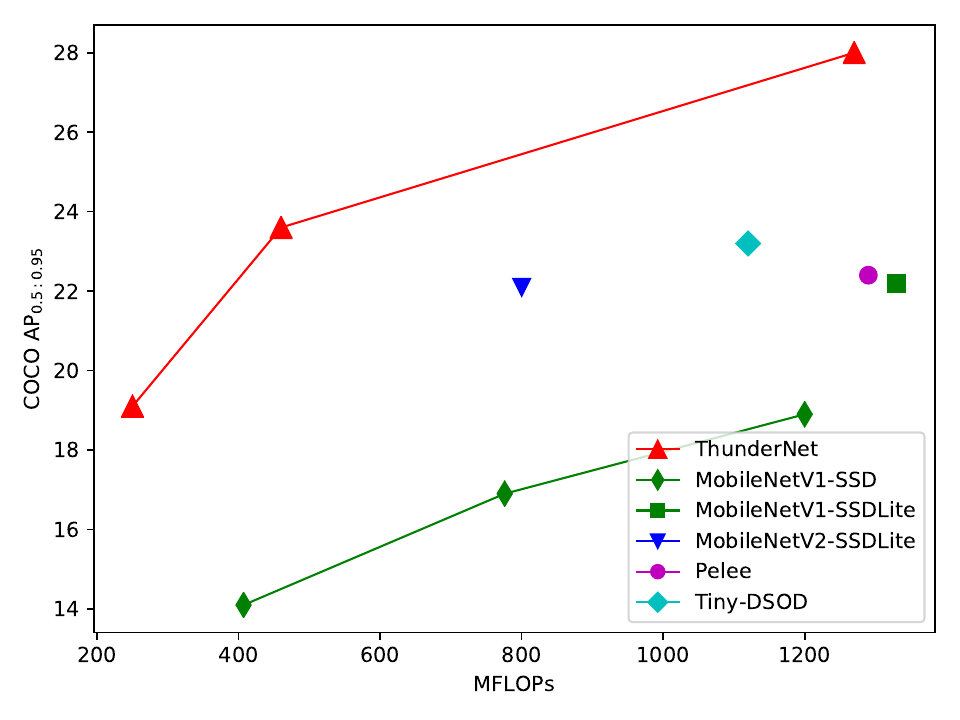}
\caption{
Comparison of ThunderNet and previous lightweight detectors on COCO test-dev.
ThunderNet achieves improvements in both accuracy and efficiency.
}
\label{figure:result-comparison}
\end{figure}


\begin{figure*}[!t]
\centering
\includegraphics[width=0.89\textwidth]{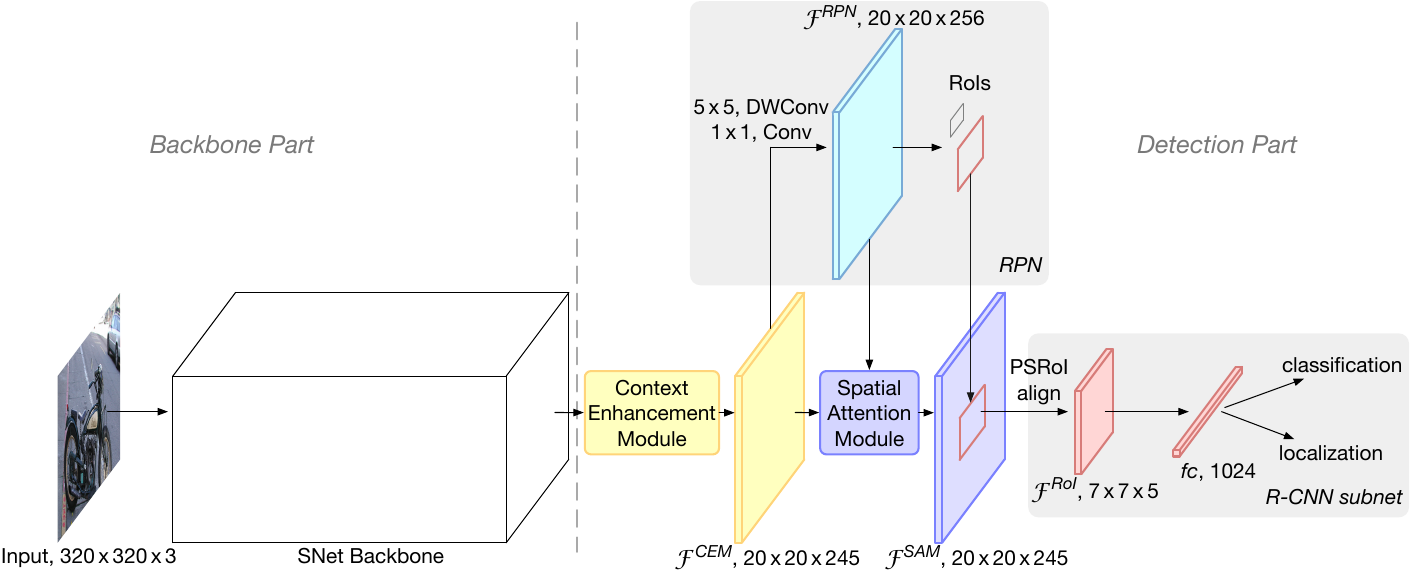}
\caption{
The overall architecture of ThunderNet.
ThunderNet uses the input resolution of 320$\times$320 pixels.
SNet backbone is based on ShuffleNetV2 and specifically designed for object detection.
In the detection part, RPN is compressed, and R-CNN subnet uses a 1024-d \emph{fc} layer for better efficiency.
Context Enhancement Module leverages semantic and context information from multiple scales.
Spatial Attention Module introduces the information from RPN to refine the feature distribution.
}
\label{figure:overall-architecture}
\end{figure*}

Real-time generic object detection on mobile devices is a crucial but challenging task in computer vision.
Compared with server-class GPUs, mobile devices are computation-constrained and raise more strict restrictions on the computational cost of detectors.
However, modern CNN-based detectors are resource-hungry and require massive computation to achieve ideal detection accuracy, which hinders them from real-time inference in mobile scenarios.

From the perspective of network structure, CNN-based detectors can be divided into the \emph{backbone part} which extracts features for the image and the \emph{detection part} which detects object instances in the image.
In the backbone part, state-of-the-art detectors are inclined to exploit huge classification networks (e.g., ResNet-101 \cite{he2016deep,dai2016r,lin2017feature,lin2017focal}) and large input images (e.g., 800$\times$1200 pixels), which requires massive computational cost.
Recent progress in lightweight image classification networks \cite{chollet2017xception,zhang2018shufflenet,ma2018shufflenet,howard2017mobilenets,sandler2018mobilenetv2} has facilitated real-time object detection \cite{howard2017mobilenets,sandler2018mobilenetv2,li2017light,ma2018shufflenet} on GPU.
However, there are several differences between image classification and object detection, e.g., object detection needs large receptive field and low-level features to improve the localization ability, which is less crucial for image classification.
The gap between the two tasks restricts the performance of these backbones on object detection and obstructs further compression without harming detection accuracy.

In the detection part, CNN-based detectors can be categorized into \emph{two-stage detectors} \cite{ren2015faster,dai2016r,lin2017feature,li2017light} and \emph{one-stage detectors} \cite{redmon2016you,liu2016ssd,redmon2017yolo9000,lin2017focal}.
For two-stage detectors, the detection part usually consists of Region Proposal Network (RPN) \cite{ren2015faster} and the detection head (including RoI warping and R-CNN subnet).
RPN first generates RoIs, and then the RoIs are further refined through the detection head.
State-of-the-art two-stage detectors tend to utilize a heavy detection part (e.g., over 10 GFLOPs \cite{ren2015faster,he2016deep,dai2016r,lin2017feature,cai2017cascade}) for better accuracy, but it is too expensive for mobile devices.
Light-Head R-CNN \cite{li2017light} adopts a lightweight detection head and achieves real-time detection on GPU.
However, when coupled with a small backbone, Light-Head R-CNN still spends more computation on the detection part than the backbone, which leads to a mismatch between a weak backbone and a strong detection part.
This imbalance not only induces great redundancy but makes the network prone to overfitting.

On the other hand, one-stage detectors directly predict bounding boxes and class probabilities.
The detection part of this category is composed of the additional layers to generate predictions, which usually involves little computation.
For this reason, one-stage detectors are widely regarded as the key to real-time detection.
However, as one-stage detectors do not conduct RoI-wise feature extraction and recognition, their results are coarser than two-stage detectors.
The problem is aggravated for lightweight detectors.
Prior lightweight one-stage detectors \cite{howard2017mobilenets,sandler2018mobilenetv2,wang2018pelee,li2018tiny} do not obtain an ideal accuracy/speed trade-off: there is a huge accuracy gap between them and the large detectors \cite{liu2016ssd,redmon2017yolo9000}, while they fail to achieve real-time detection on mobile devices.
It inspires us to rethink: \emph{can two-stage detectors surpass one-stage detectors in real-time detection?}

In this paper, we propose a lightweight two-stage generic object detector named \emph{ThunderNet}.
The design of ThunderNet aims at the computationally expensive structures in state-of-the-art two-stage detectors.
In the backbone part, we investigate the drawbacks in previous lightweight backbones, and present a lightweight backbone named \emph{SNet} designed for object detection.
In the detection part, we follow the detection head design in Light-Head R-CNN, and further compress RPN and R-CNN subnet.
To eliminate the performance degradation induced by small backbones and small feature maps, we design two efficient architecture blocks, \emph{Context Enhancement Module} (CEM) and \emph{Spatial Attention Module} (SAM).
CEM combines the feature maps from multiple scales to leverage local and global context information, while SAM uses the information learned in RPN to refine the feature distribution in RoI warping.
At last, we investigate the balance between the input resolution, the backbone, and the detection head.
Fig.~\ref{figure:overall-architecture} illustrates the overall architecture of ThunderNet.

ThunderNet surpasses prior lightweight one-stage detectors with significantly less computational cost on PASCAL VOC \cite{everingham2010pascal} and COCO \cite{lin2014microsoft} benchmarks (Fig.~\ref{figure:result-comparison}).
ThunderNet outperforms Tiny-DSOD \cite{li2018tiny} with only 42\% of the computational cost and obtains gains of 6.5 mAP on VOC and 4.8 AP on COCO under similar complexity.
Without bells and whistles, ThunderNet runs in \emph{real time} on ARM (24.1 fps) and x86 (47.3 fps) with MobileNet-SSD level accuracy.
To the best of our knowledge, this is the \emph{first} real-time detector and the \emph{fastest} single-thread speed reported on ARM platforms.
\emph{These results have demonstrated the effectiveness of two-stage detectors in real-time object detection.}


\section{Related Work}

\head{CNN-based object detectors.}
CNN-based object detectors are commonly classified into two-stage detectors and one-stage detectors.
In two-stage detectors, R-CNN \cite{girshick2014rich} is among the earliest CNN-based detection systems.
Since then, progressive improvements \cite{he2014spatial,girshick2015fast} are proposed for better accuracy and efficiency.
Faster R-CNN \cite{ren2015faster} proposes Region Proposal Network (RPN) to generate regions proposals instead of pre-handled proposals.
R-FCN \cite{dai2016r} designs a fully convolutional architecture which shares computation on the entire image.
On the other hand, one-stage detectors such as SSD \cite{liu2016ssd} and YOLO \cite{redmon2016you,redmon2017yolo9000,redmon2018yolov3} achieve real-time inference on GPU with very competitive accuracy.
RetinaNet \cite{lin2017focal} proposes focal loss to address the foreground-background class imbalance and achieves significant accuracy improvements.
In this work, we present a two-stage detector which focuses on efficiency.

\head{Real-time generic object detection.}
Real-time object detection is another important problem for CNN-based detectors.
Commonly, one-stage detectors are regarded as the key to real-time detection.
For instance, YOLO \cite{redmon2016you,redmon2017yolo9000,redmon2018yolov3} and SSD \cite{liu2016ssd} run in real time on GPU.
When coupled with small backbone networks, lightweight one-stage detectors, such as MobileNet-SSD \cite{howard2017mobilenets}, MobileNetV2-SSDLite \cite{sandler2018mobilenetv2}, Pelee \cite{wang2018pelee} and Tiny-DSOD \cite{li2018tiny}, achieve inference on mobile devices at low frame rates.
For two-stage detectors, Light-Head R-CNN \cite{li2017light} utilizes a light detection head and runs at over 100 fps on GPU.
This raises a question: are two-stage detectors better than one-stage detectors in real-time detection?
In this paper, we present the effectiveness of two-stage detectors in real-time detection.
Compared with prior lightweight one-stage detectors, ThunderNet achieves a better balance between accuracy and efficiency.

\head{Backbone networks for detection.}
Modern CNN-based detectors typically adopt image classification networks \cite{simonyan2014very,he2016deep,xie2017aggregated,hu2018squeeze} as the backbones.
FPN \cite{lin2017feature} exploits the inherent multi-scale, pyramidal hierarchy of CNNs to construct feature pyramids.
Lightweight detectors also benefit from the recent progress in small networks, such as MobileNet \cite{howard2017mobilenets,sandler2018mobilenetv2} and ShuffleNet \cite{zhang2018shufflenet,ma2018shufflenet}.
However, image classification and object detection require different properties of networks.
Therefore, simply transferring classification networks to object detection is not optimal.
For this reason, DetNet \cite{li2018detnet} designs a backbone specifically for object detection.
Recent lightweight detectors \cite{wang2018pelee,li2018tiny} also design specialized backbones.
However, this area is still not well studied.
In this work, we investigate the drawbacks of prior lightweight backbones and present a lightweight backbone for real-time detection task.
  

\section{ThunderNet}

In this section, we present the details of ThunderNet.
Our design mainly focuses on efficiency, but our model still achieves superior accuracy.

\subsection{Backbone Part}

\paragraph{Input Resolution.}

The input resolution of two-stage detectors is usually very large, e.g., FPN \cite{lin2017feature} uses input images of 800$\times$ pixels.
It brings several advantages but involves enormous computational cost as well.
To improve the inference speed, ThunderNet utilizes the input resolution of 320$\times$320 pixels.
Moreover, in practice, we observe that \emph{the input resolution should match the capability of the backbone}.
A small backbone with large inputs and a large backbone with small inputs are both not optimal.
Details are discussed in Sec.~\ref{section:ablation-study-input-resolution}.

\vspace{-12pt}
\paragraph{Backbone Networks.}

\begin{table}[t]
\scriptsize
\centering
\begin{tabular}{l|c|c|c|c}
\multirow{2}{*}{Stage} & \multirow{2}{*}{\begin{tabular}[c]{@{}c@{}}Output\\ Size\end{tabular}} & \multicolumn{3}{c}{Layer} \\ \cline{3-5}
 & & SNet49 & SNet146 & SNet535 \\ \hlineB{2.5}
Input & 224$\times$224 & \multicolumn{3}{c}{image} \\ \hline
Conv1 & 112$\times$112 & 3$\times$3, 24, s2 & 3$\times$3, 24, s2 & 3$\times$3, 48, s2 \\ \hline
Pool & 56$\times$56 & \multicolumn{3}{c}{3$\times$3 maxpool, s2} \\ \hline
\multirow{2}{*}{Stage2} & 28$\times$28 & [60, s2] & [132, s2] & [248, s2] \\
 & 28$\times$28 & [60, s1]$\times$3 & [132, s1]$\times$3 & [248, s1]$\times$3 \\ \hline
\multirow{2}{*}{Stage3} & 14$\times$14 & [120, s2] & [264, s2] & [496, s2] \\
 & 14$\times$14 & [120, s1]$\times$7 & [264, s1]$\times$7 & [496, s1] $\times$7 \\ \hline
\multirow{2}{*}{Stage4} & 7$\times$7 & [240, s2] & [528, s2] & [992, s2] \\
 & 7$\times$7 & [240, s1]$\times$3 & [528, s1]$\times$3 & [992, s1]$\times$3 \\ \hline
Conv5 & 7$\times$7 & 1$\times$1, 512 & - & - \\ \hline
Pool & 1$\times$1 & \multicolumn{3}{c}{global avg pool} \\ \hline
FC &  & \multicolumn{3}{c}{1000-d fc} \\ \hline
FLOPs &  & 49M & 146M & 535M \\
\end{tabular}
\vspace{3pt}
\caption{
Architecture of the SNet backbone networks.
SNet uses ShuffleNetV2 basic blocks but replaces all 3$\times$3 depthwise convolutions with 5$\times$5 depthwise convolutions.
}
\label{table:architecture-backbone}
\end{table}

Backbone networks provide basic feature representation of the input image and have great influence on both accuracy and efficiency.
CNN-based detectors usually use classification networks transferred from ImageNet classification as the backbone.
However, as image classification and object detection require different properties from the backbone, simply transferring classification networks to object detection is not optimal.

\emph{Receptive field}:
The receptive field size plays an important role in CNN models.
CNNs can only capture information inside the receptive field.
Thus, a large receptive field can leverage more context information and encode long-range relationship between pixels more effectively.
This is crucial for the localization subtask, especially for the localization of large objects.
Previous works \cite{peng2017large,li2017light} have also demonstrated the effectiveness of the large receptive field in semantic segmentation and object detection.

\emph{Early-stage and late-stage features}:
In the backbone, early-stage feature maps are larger with low-level features which describe spatial details, while late-stage feature maps are smaller with high-level features which are more discriminative.
Generally, localization is sensitive to low-level features while high-level features are crucial for classification.
In practice, we observe that \emph{localization is more difficult than classification for larger backbones}, which indicates that early-stage features are more important.
And \emph{the weak representation power restricts the accuracy in both subtasks for extremely tiny backbones}, suggesting that both early-stage and late-stage features are crucial at this level.

The designs of prior lightweight backbones violate the aforementioned factors: ShuffleNetV1/V2 \cite{zhang2018shufflenet,ma2018shufflenet} have restricted receptive field (121 pixels vs. 320 pixels of input), ShuffleNetV2 \cite{ma2018shufflenet} and MobileNetV2 \cite{sandler2018mobilenetv2} lack early-stage features, and Xception \cite{chollet2017xception} suffer from the insufficient high-level features under small computational budgets.

Based on these insights, we start from ShuffleNetV2, and build a lightweight backbone named \emph{SNet} for real-time detection.
We present three SNet backbones: \emph{SNet49} for faster inference, \emph{SNet535} for better accuracy, and \emph{SNet146} for a better speed/accuracy trade-off.
First, we replace all 3$\times$3 depthwise convolutions in ShuffleNetV2 with 5$\times$5 depthwise convolutions.
In practice, 5$\times$5 depthwise convolutions provide similar runtime speed to 3$\times$3 counterparts while effectively enlarging the receptive field (from 121 to 193 pixels).
In SNet146 and SNet535, we remove Conv5 and add more channels in early stages.
This design generates more low-level features without additional computational cost.
In SNet49, we compress Conv5 to 512 channels instead of removing it and increase the channels in the early stages for a better balance between low-level and high-level features.
If we remove Conv5, the backbone cannot encode adequate information.
But if the 1024-d Conv5 layer is preserved, the backbone suffers from limited low-level features.
Table~\ref{table:architecture-backbone} shows the overall architecture of the backbones.
Besides, the last output feature maps of Stage3 and Stage4 (Conv5 for SNet49) are denoted as $C_4$ and $C_5$.

\subsection{Detection Part}

\paragraph{Compressing RPN and Detection Head.}

Two-stage detectors usually adopt large RPN and a heavy detection head.
Although Light-Head R-CNN \cite{li2017light} uses a lightweight detection head, it is still too heavy when coupled with small backbones and induces imbalance between the backbone and the detection part.
This imbalance not only leads to redundant computation but increases the risk of overfitting.

To address this issue, we compress RPN by replacing the original 256-channel 3$\times$3 convolution with a 5$\times$5 depthwise convolution and a 256-channel 1$\times$1 convolution.
We increase the kernel size to enlarge the receptive field and encode more context information.
Five scales \{$32^2$, $64^2$, $128^2$, $256^2$, $512^2$\} and five aspect ratios \{1:2, 3:4, 1:1, 4:3, 2:1\} are used to generate anchor boxes.
Other hyperparameters remain the same as in \cite{li2017light}.

In the detection head, Light-Head R-CNN generates a thin feature map with $\alpha \times p \times p$ channels before RoI warping, where $p=7$ is the pooling size and $\alpha=10$.
As the backbones and the input images are smaller in ThunderNet, we further narrow the feature map by halving $\alpha$ to 5 to eliminate redundant computation.
For RoI warping, we opt for PSRoI align as it squeezes the number of channels to $\alpha$.

As the RoI feature from PSRoI align is merely 245-d, we apply a 1024-d fully-connected (\emph{fc}) layer in R-CNN subnet.
As demonstrated in Sec.~\ref{section:ablation-study-detection-part}, this design further reduces the computational cost of R-CNN subnet without sacrificing accuracy.
Besides, due to the small feature maps, we reduce the number of RoIs for testing as discussed in Sec.~\ref{section:implementation-details}.

\vspace{-12pt}
\paragraph{Context Enhancement Module.}


\begin{figure}[t]
\centering
\includegraphics[width=0.45\textwidth]{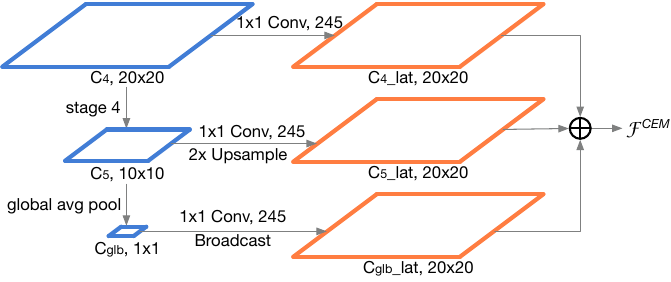}
\caption{
Structure of Context Enhancement Module (CEM).
CEM combines feature maps from three scales and encodes more context information.
It enlarges the receptive field and generates more discriminative features.
}
\label{figure:context-enhancemene-module}
\end{figure}

Light-Head R-CNN applies Global Convolutional Network (GCN) \cite{peng2017large} to generate the thin feature map.
It significantly increases the receptive field but involves enormous computational cost.
Coupled with SNet146, GCN requires 2$\times$ the FLOPs needed by the backbone (596M vs. 298M).
For this reason, we decide to abandon this design in ThunderNet.

However, the network suffers from the small receptive field and fails to encode sufficient context information without GCN.
A common technique to address this issue is Feature Pyramid Network (FPN) \cite{lin2017feature}.
However, prior FPN structures \cite{lin2017feature,fu2017dssd,li2018tiny,redmon2018yolov3} involve many extra convolutions and multiple detection branches, which increases the computational cost and induces enormous runtime latency.

For this reason, we design an efficient \emph{Context Enhancement Module} (CEM) to enlarge the receptive field.
The key idea of CEM is to aggregate multi-scale \emph{local} context information and \emph{global} context information to generate more discriminative features.
In CEM, the feature maps from three scales are merged: $C_4$, $C_5$ and $C_{glb}$.
$C_{glb}$ is the global context feature vector by applying a global average pooling on $C_5$.
We then apply a $1 \times 1$ convolution on each feature map to squeeze the number of channels to $\alpha \times p \times p = 245$.
Afterwards, $C_5$ is upsampled by $2 \times$ and $C_{glb}$ is broadcast so that the spatial dimensions of the three feature maps are equal.
At last, the three generated feature maps are aggregated.
By leveraging both local and global context, CEM effectively enlarges the receptive field and refines the representation ability of the thin feature map.
Compared with prior FPN structures, CEM involves only two 1$\times$1 convolutions and a \emph{fc} layer, which is more computation-friendly.
Fig.~\ref{figure:context-enhancemene-module} illustrates the structure of this module.

\vspace{-12pt}
\paragraph{Spatial Attention Module.}

During RoI warping, we expect the features in the background regions to be small and the foreground counterparts to be high.
However, compared with large models, as ThunderNet utilizes lightweight backbones and small input images, it is more difficult for the network itself to learn a proper feature distribution.

\begin{figure}[!t]
\centering
\includegraphics[width=0.45\textwidth]{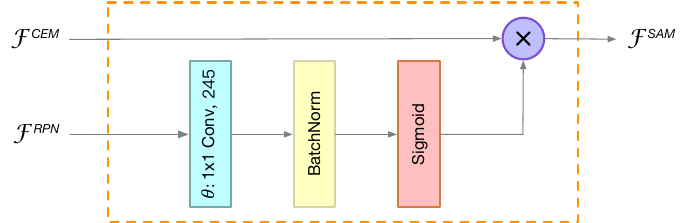}
\caption{
Structure of Spatial Attention Module (SAM).
SAM leverages the information learned in RPN to refine the feature distribution of the feature map from Context Enhancement Module.
The feature map is then used for RoI warping.
}
\label{figure:spatial-attention-module}
\end{figure}

For this reason, we design a computation-friendly \emph{Spatial Attention Module} (SAM) to explicitly re-weight the feature map before RoI warping over the spatial dimensions.
The key idea of SAM is to use the knowledge from RPN to refine the feature distribution of the feature map.
RPN is trained to recognize foreground regions under the supervision of ground truths.
Therefore, the intermediate features in RPN can be used to distinguish foreground features from background features.
SAM accepts two inputs: the intermediate feature map from RPN $\mathcal{F}^{\textit{RPN}}$ and the thin feature map from CEM $\mathcal{F}^{\textit{CEM}}$.
The output of SAM $\mathcal{F}^{\textit{SAM}}$ is defined as:
\begin{equation}
\mathcal{F}^{\textit{SAM}} = \mathcal{F}^{\textit{CEM}} \cdot sigmoid(\theta(\mathcal{F}^{\textit{RPN}})).
\end{equation}
Here $\theta(\cdot)$ is a dimension transformation to match the number of channels in both feature maps.
The sigmoid function is used to constrain the values within [0, 1].
At last, $\mathcal{F}^{\textit{CEM}}$ is re-weighted by the generated feature map for better feature distribution.
For computational efficiency, we simply apply a 1$\times$1 convolution as $\theta(\cdot)$, so the computational cost of CEM is negligible.
Fig.~\ref{figure:spatial-attention-module} shows the structure of SAM.


SAM has two functions.
The first one is to refine the feature distribution by strengthening foreground features and suppressing background features.
The second one is to stabilize the training of RPN as SAM enables extra gradient flow from R-CNN subnet to RPN:
\begin{equation}
\frac{\partial \mathcal{L}}{\partial \mathcal{F}^{\textit{RPN}}_i} = \frac{\partial \mathcal{L}^{\textit{RPN}}}{\partial \mathcal{F}^{\textit{RPN}}_i} + \sum\nolimits_{\forall j}\frac{\partial \mathcal{L}^{\textit{R-CNN}}}{\partial \mathcal{F}^{\textit{SAM}}_j} \cdot \frac{\partial \mathcal{F}^{\textit{SAM}}_j}{\partial \mathcal{F}^{\textit{RPN}}_i}.
\end{equation}
As a result, RPN receives additional supervision from R-CNN subnet, which helps the training of RPN.


\section{Experiments}

In this section, we evaluate the effectiveness of ThunderNet on PASCAL VOC \cite{everingham2010pascal} and COCO \cite{lin2014microsoft} benchmarks.
Then we conduct ablation studies to evaluate our design.

\begin{table*}[!t]
\scriptsize
\centering
\begin{tabular}{p{11em}|P{10em}|P{8em}|P{6em}|P{4em}}
Model & Backbone & Input & MFLOPs & mAP \\ \hlineB{2.5}
YOLOv2 \cite{redmon2017yolo9000} & Darknet-19 & $416 \times 416$ & 17400 & 76.8 \\
SSD300* \cite{liu2016ssd} & VGG-16 & $300 \times 300$ & 31750 & 77.5 \\
SSD321 \cite{fu2017dssd} & ResNet-101 & $321 \times 321$ & 15400 & 77.1 \\
DSSD321 \cite{fu2017dssd} & ResNet-101 + FPN & $321 \times 321$ & 21200 & 78.6 \\
R-FCN \cite{dai2016r} & ResNet-50 & $600 \times 1000$ & 58900 & 77.4 \\ \hline
Tiny-YOLO \cite{redmon2017yolo9000} & Tiny Darknet & $416 \times 416$ & 3490 & 57.1 \\
D-YOLO \cite{mehta2018object} & Tiny Darknet & $416 \times 416$ & 2090 & 67.6 \\
MobileNet-SSD \cite{wang2018pelee} & MobileNet & $300 \times 300$ & 1150 & 68.0 \\
Pelee \cite{wang2018pelee} & PeleeNet & $304 \times 304$ & 1210 & 70.9 \\
Tiny-DSOD \cite{li2018tiny} & DDB-Net + D-FPN & $300 \times 300$ & 1060 & 72.1 \\ \hline
ThunderNet (\emph{ours}) & SNet49 & $320 \times 320$ & \textbf{250} & 70.1 \\
ThunderNet (\emph{ours}) & SNet146 & $320 \times 320$ & \textbf{461} & \textbf{75.1} \\
ThunderNet (\emph{ours}) & SNet535 & $320 \times 320$ & 1287 & \textbf{78.6} \\
\end{tabular}
\vspace{3pt}
\caption{
Evaluation results on VOC 2007 test.
ThunderNet surpasses competing models with significantly less computational cost.
}
\label{table:results-voc}
\end{table*}

\begin{table*}[!t]
\setlength{\tabcolsep}{10pt}
\centering
\scriptsize
\begin{tabular}{l|c|c|c|ccc}
Model & Backbone & Input & MFLOPs & AP & AP$_{50}$ & AP$_{75}$ \\ \hlineB{2.5}
YOLOv2 \cite{redmon2017yolo9000} & Darknet-19 & $416 \times 416$ & 17500 & 21.6 & 44.0 & 19.2 \\
SSD300* \cite{liu2016ssd} & VGG-16 & $300 \times 300$ & 35200 & 25.1 & 43.1 & 25.8 \\ 
SSD321 \cite{fu2017dssd} & ResNet-101 & $321 \times 321$ & 16700 & 28.0 & 45.4 & 29.3 \\
DSSD321 \cite{fu2017dssd} & ResNet-101 + FPN & $321 \times 321$ & 22300 & 28.0 & 46.1 & 29.2 \\
Light-Head R-CNN \cite{ma2018shufflenet} & ShuffleNetV2* & $800 \times 1200$ & 5650 & 23.7 & - & - \\ \hline
MobileNet-SSD \cite{howard2017mobilenets} & MobileNet & $300 \times 300$ & 1200 & 19.3 & - & - \\
MobileNet-SSDLite \cite{sandler2018mobilenetv2} & MobileNet & $320 \times 320$ & 1300 & 22.2 & - & - \\
MobileNetV2-SSDLite \cite{sandler2018mobilenetv2} & MobileNetV2 & $320 \times 320$ & 800 & 22.1 & - & - \\
Pelee \cite{wang2018pelee} & PeleeNet & $304 \times 304$ & 1290 & 22.4 & 38.3 & 22.9 \\
Tiny-DSOD \cite{li2018tiny} & DDB-Net + D-FPN & $300 \times 300$ & 1120 & 23.2 & 40.4 & 22.8 \\ \hline
ThunderNet (\emph{ours}) & SNet49 & $320 \times 320$ & \textbf{262} & 19.2 & 33.7 & 19.7 \\
ThunderNet (\emph{ours}) & SNet146 & $320 \times 320$ & \textbf{473} & \textbf{23.7} & 40.3 & \textbf{24.6} \\
ThunderNet (\emph{ours}) & SNet535 & $320 \times 320$ & 1300 & \textbf{28.1} & \textbf{46.2} & \textbf{29.6} \\
\end{tabular}
\vspace{3pt}
\caption{
Evaluation results on COCO test-dev.
ThunderNet with SNet49 achieves MobileNet-SSD level accuracy with 22\% of the FLOPs.
ThunderNet with SNet146 achieves superior accuracy to prior lightweight one-stage detectors with merely 40\% of the FLOPs.
ThunderNet with SNet535 rivals large detectors with significantly less computational cost.
}
\label{table:results-coco}
\end{table*}

\subsection{Implementation Details}
\label{section:implementation-details}

Our detectors are trained end-to-end on 4 GPUs using synchronized SGD with a weight decay of 0.0001 and a momentum of 0.9.
The batch size is set to 16 images per GPU.
Each image has 2000/200 RoIs for training/testing.
For efficiency, the input resolution of 320$\times$320 pixels is used instead of 600$\times$ or 800$\times$ pixels in common large two-stage detectors.
Multi-scale training with \{240, 320, 480\} pixels is adopted.
As the input resolution is small, we use heavy data augmentation \cite{liu2016ssd}.
The networks are trained for 62.5K iterations on VOC dataset and 375K iterations on COCO dataset.
The learning rate starts from 0.01 and decays by a factor of 0.1 at 50\% and 75\% of the total iterations.
Online hard example mining \cite{shrivastava2016training} is adopted and Soft-NMS \cite{bodla2017soft} is used for post-processing.
Cross-GPU Batch Normalization (CGBN) \cite{peng2018megdet} is used to learn batch normalization statistics.

\subsection{Results on PASCAL VOC}

PASCAL VOC dataset consists of natural images drawn from 20 classes.
The networks are trained on the union set of VOC 2007 trainval and VOC 2012 trainval, and we report single-model results on VOC 2007 test.
The results are exhibited in Table~\ref{table:results-voc}.

ThunderNet surpasses prior state-of-the-art lightweight one-stage detectors.
ThunderNet with SNet49 outperforms MobileNet-SSD with merely 21\% of the FLOPs, while the SNet146-based model surpasses Tiny-DSOD by 2.9 mAP with about 43\% of the FLOPs.
Moreover, ThunderNet with SNet146 performs better than Tiny-DSOD by 6.5 mAP under similar computational cost.

Furthermore, ThunderNet achieves superior results to state-of-the-art large object detectors such as YOLOv2 \cite{redmon2017yolo9000}, SSD300* \cite{liu2016ssd}, SSD321 \cite{liu2016ssd} and R-FCN \cite{dai2016r}, and is on a par with DSSD321 \cite{fu2017dssd}, but reduces the computational cost by orders of magnitude.
We note that the backbone of ThunderNet is significantly weaker and smaller than the large detectors.
It demonstrates that ThunderNet achieves a much better trade-off between accuracy and efficiency.

\subsection{Results on MS COCO}

MS COCO dataset consists of natural images from 80 object categories.
Following common practice \cite{lin2017feature,li2017light}, we use trainval35k for training, minival for validation, and report single-model results on test-dev.

As shown in Table~\ref{table:results-coco}, ThunderNet with SNet49 achieves MobileNet-SSD level accuracy with 22\% of the FLOPs.
ThunderNet with SNet146 surpasses MobileNet-SSD \cite{howard2017mobilenets}, MobileNet-SSDLite \cite{sandler2018mobilenetv2}, and Pelee \cite{wang2018pelee} with less than 40\% of the computational cost.
It is noteworthy that our approach achieves considerably better AP$_{75}$, which suggests our model performs better in localization.
This is consistent with our initial motivation to design two-stage real-time detectors.
Compared with Tiny-DSOD \cite{li2018tiny}, ThunderNet achieves better AP but worse AP$_{50}$ with 42\% of the FLOPs.
We conjecture that deep supervision and feature pyramid in Tiny-DSOD contribute to better classification accuracy.
However, ThunderNet is still better in localization.

\begin{figure}[!t]
\centering
\includegraphics[width=0.47\textwidth]{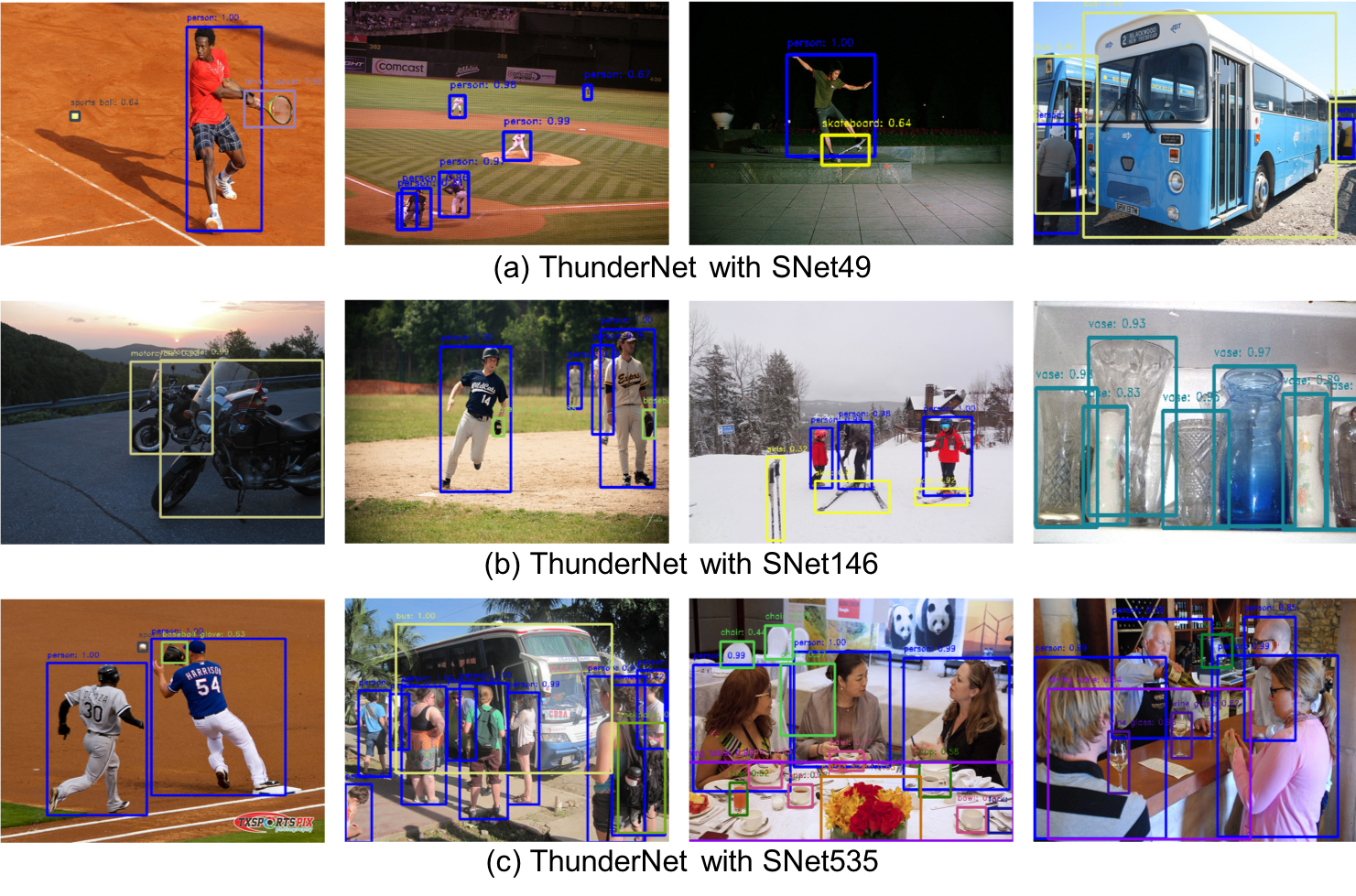}
\caption{Examples visualization on COCO test-dev.}
\label{figure:coco-visualization}
\end{figure}

ThunderNet with SNet535 achieves significantly better detection accuracy under comparable computational cost.
As shown in Table~\ref{table:results-coco}, ThunderNet surpasses other one-stage counterparts by at least 4.8 AP, 5.8 AP$_{50}$ and 6.7 AP$_{75}$.
The gap in AP$_{75}$ is larger than the gap in AP$_{50}$, which means our model provides more accurate bounding boxes than other detectors.
This further demonstrates that two-stage detectors are prior to one-stage detectors in real-time detection task.
Fig.~\ref{figure:coco-visualization} visualizes several examples on COCO test-dev.


We also compare ThunderNet with large one-stage detectors.
ThunderNet with SNet146 surpasses YOLOv2 \cite{redmon2017yolo9000} with 37$\times$ fewer FLOPs.
And ThunderNet with SNet535 significantly outperforms YOLOv2 and SSD300 \cite{liu2016ssd}, and rivals SSD321 \cite{fu2017dssd} and DSSD321 \cite{fu2017dssd}.
It suggests that ThunderNet is not only efficient but highly accurate.

\subsection{Ablation Experiments}

\subsubsection{Input Resolution}
\label{section:ablation-study-input-resolution}

We first explore the relationship between the input resolution and the backbone.
Table~\ref{table:input-resolution} reveals that large backbones with small images and small backbones with large images are both not optimal.
There is a trade-off between the two factors.
On the one hand, small images lead to low-resolution feature maps and induce severe loss of detail features.
It is hard to be remedied by simply increasing the capacity of the backbones.
On the other hand, small backbones are too weak to encode sufficient information from large images.
The backbone and the input images should match for a better balance between the representation ability and the resolution of the feature maps.

\vspace{-5pt}
\subsubsection{Backbone Networks}
\label{section:ablation-study-backbone-networks}

We then evaluate the design of the backbones.
SNet146 and SNet49 are used as the baselines.
SNet146 achieves 32.5\% top-1 error on ImageNet classification and 23.6 AP on COCO test-dev (Table~\ref{table:ablation-study-backbone-networks}(a)), while SNet49 achieves 39.7\% top-1 error and 19.1 AP (Table~\ref{table:ablation-study-backbone-networks}(e)).

\vspace{-12pt}
\paragraph{5$\times$5 Depthwise Convolutions.}

We evaluate the effectiveness of 5$\times$5 depthwise convolutions on SNet146.
We first replace all 5$\times$5 depthwise convolutions with 3$\times$3 depthwise convolutions.
For fair comparison, the channels from Stage2 to Stage4 are slightly increased to maintain the computational cost unchanged.
This model performs worse on both image classification (by 0.2\%) and object detection (by 0.9 AP) (Table~\ref{table:ablation-study-backbone-networks}(b)).
Compared with 3$\times$3 depthwise convolutions, 5$\times$5 depthwise convolutions considerably increase the receptive fields, which helps in both tasks.

We then add another 3$\times$3 depthwise convolution before the first 1$\times$1 convolution in all building blocks as in ShuffleNetV2* \cite{ma2018shufflenet}.
The number of channels is kept unchanged as the baseline.
This model is comparable on image classification, but slightly worse on object detection (by 0.3 AP) (Table~\ref{table:ablation-study-backbone-networks}(c)).
As this model and SNet146 have the same receptive fields theoretically, we conjecture that 5$\times$5 depthwise convolutions can provide larger valid receptive fields, which is especially crucial in object detection.

\vspace{-12pt}
\paragraph{Early-stage and Late-stage Features.}

To investigate the trade-off between early-stage and late-stage features, we first add a 1024-channel Conv5 in SNet146.
The channels in the early stages are reduced accordingly.
This model slightly improves the top-1 error, but reduces AP by 0.4 (Table~\ref{table:ablation-study-backbone-networks}(d)).
A wide Conv5 generates more discriminative features, which improves the classification accuracy.
However, object detection focuses on both classification and localization.
Increasing the channels in early stages encodes more detail information, which is beneficial for localization.

For SNet49, we first remove Conv5 in SNet49 and increase the channels from Stage2 to Stage4.
Table~\ref{table:ablation-study-backbone-networks}(f) shows that both the classification and the detection performance suffer from severe degradation.
Removing Conv5 cuts the output channels of the backbone by half, which hinders the model from learning adequate information.

We then extend Conv5 to 1024 channels as in the original ShuffleNetV2.
The early-stage channels are compressed to maintain the same overall computational cost.
This model surpasses SNet49 on image classification by 0.8\%, but performs worse on object detection (Table~\ref{table:ablation-study-backbone-networks}(g)).
By leveraging a wide Conv5, this model benefits from more high-level features in image classification.
However, it suffers from the lack of low-level features in object detection.
It further demonstrates the differences between image classification and object detection.

\begin{table}[!t]
\setlength{\tabcolsep}{10pt}
\centering
\scriptsize
\begin{tabular}{l|c|c|P{3em}}
Backbone & Input & MFLOPs & AP \\ \hlineB{2.5}
SNet49 & $320 \times 320$ & 262 & \textbf{19.2} \\
SNet146 & $224 \times 224$ & 267 & 18.7 \\
SNet535 & $128 \times 128$ & 265 & 13.2 \\ \hline
SNet49 & $480 \times 480$ & 506 & 22.0 \\
SNet146 & $320 \times 320$ & 473 & \textbf{23.7} \\
SNet535 & $192 \times 192$ & 512 & 20.2 \\
\end{tabular}
\vspace{3pt}
\caption{Evaluation of different input resolutions on COCO test-dev. Large backbones with small images and small backbones with large images are both not optimal.}
\label{table:input-resolution}
\end{table}

\begin{table}[!t]
\scriptsize
\centering
\begin{tabular}{c@{\hskip 2pt}l|c|c|P{3em}}
\multicolumn{2}{l|}{Backbone} & MFLOPs & Top-1 Err. & AP \\ \hlineB{2.5}
(a) & SNet146 & 146 & 32.5 & \textbf{23.7} \\
(b) & SNet146 + 3$\times$3 DWConv & 145 & 32.7 & 22.7 \\
(c) & SNet146 + double 3$\times$3 DWConv & 143 & 32.4 & 23.3 \\
(d) & SNet146 + 1024-d Conv5 & 147 & \textbf{32.3} & 23.2 \\ \hline
(e) & SNet49 & 49 & 39.7 & \textbf{19.2} \\
(f) & SNet49 + No Conv5 & 49 & 40.8 & 18.2 \\
(g) & SNet49 + 1024-d Conv5 & 49 & \textbf{38.9} & 18.8 \\
\end{tabular}
\vspace{3pt}
\caption{
Evaluation of different backbones on ImageNet classification and COCO test-dev.
\textbf{DWConv}: depthwise convolution.
}
\label{table:ablation-study-backbone-networks}
\end{table}

\begin{table}[!t]
\setlength{\tabcolsep}{10pt}
\centering
\scriptsize
\begin{tabular}{l|c|c|P{3em}}
Backbone & MFLOPs & Top-1 Err. & AP \\ \hlineB{2.5}
ShuffleNetV1 \cite{zhang2018shufflenet} & 137 & 34.8 & 20.8 \\
ShuffleNetV2 \cite{ma2018shufflenet} & 147 & \textbf{31.4} & 22.7 \\
ShuffleNetV2* \cite{ma2018shufflenet} & 145 & 32.2 & 23.2 \\
Xception \cite{chollet2017xception} & 145 & 34.1 & 23.0 \\
MobileNetV2 \cite{sandler2018mobilenetv2} & 145 & 32.9 & 22.7 \\ \hline
SNet146 & 146 & 32.5 & \textbf{23.7} \\
\end{tabular}
\vspace{3pt}
\caption{Evaluation of lightweight backbones on COCO test-dev.
SNet146 achieves better detection results though the classification accuracy is lower.}
\label{table:backbones}
\end{table}

\vspace{-12pt}
\paragraph{Comparison with Lightweight Backbones.}

We further compare SNet with other lightweight backbones in ThunderNet framework (Table~\ref{table:backbones}).
SNet146 surpasses Xception \cite{chollet2017xception}, MobileNetV2 \cite{sandler2018mobilenetv2}, and ShuffleNetV1/V2/V2* \cite{zhang2018shufflenet,ma2018shufflenet} on object detection under similar FLOPs.
These results further demonstrate the effectiveness of our design.

\vspace{-5pt}
\subsubsection{Detection Part}
\label{section:ablation-study-detection-part}

We also investigate the effectiveness of the design of the detection part in ThunderNet.
Table~\ref{table:ablation-study-architecture-modules} describes the comparison of the model variants in the experiments.

\vspace{-12pt}
\paragraph{Baseline.}

We choose a compressed Light-Head R-CNN \cite{li2017light} with SNet146 as the baseline.
$C_5$ is upsampled by $2\times$ to obtain the same downsampling rate.
$C_4$ and $C_5$ are then squeezed to 245 channels and sent to RPN and RoI warping respectively.
We use a 256-channel 3$\times$3 convolution in RPN and a 2048-d \emph{fc} layer in R-CNN subnet.
This model requires 703 MFLOPs and achieves 21.9 AP (Table~\ref{table:ablation-study-architecture-modules}(a)).
Besides, we would mention that multi-scale training, CGBN \cite{peng2018megdet}, and Soft-NMS \cite{bodla2017soft} gradually improve the baseline by 1.4 AP (from 20.5 to 21.9 AP).

\vspace{-12pt}
\paragraph{RPN and R-CNN subnet.}

We first replace the 3$\times$3 convolution in RPN with a 5$\times$5 depthwise convolution and a 1$\times$1 convolution.
The number of output channels remains unchanged.
This design reduces the computational cost by 28\% without harming the accuracy (Table~\ref{table:ablation-study-architecture-modules}(b)).
We then halve the number of outputs of the \emph{fc} layer in R-CNN subnet to 1024, which achieves a further 13\% compression on the FLOPs with a marginal decrease of 0.2 AP. (Table~\ref{table:ablation-study-architecture-modules}(c)).
These results demonstrate that heavy RPN and R-CNN subnet introduce great redundancy for lightweight detectors.
More details will be discussed in Sec.~\ref{section:large-backbone-or-heavy-head}.

\vspace{-12pt}
\paragraph{Context Enhancement Module.}

\begin{table}[!t]
\scriptsize
\setlength{\tabcolsep}{4pt}
\centering
\begin{tabular}{cccccc|ccc|c}
 & BL & SRPN & SRCN & CEM & SAM & AP & AP$_{50}$ & AP$_{75}$ & MFLOPs \\ \hlineB{2.5}
(a) & \checkmark & & & & & 21.9 & 37.6 & 22.5 & 714 \\
(b) & & \checkmark & & & & 21.8 & 37.5 & 22.4 & 516 \\
(c) & & \checkmark & \checkmark & & & 21.6 & 37.4 & 22.2 & \textbf{448} \\
(d) & & \checkmark & \checkmark & \checkmark & & 23.3 & 39.9 & 24.0 & 449 \\
(e) & & \checkmark & \checkmark & & \checkmark & 23.0 & 39.0 & 24.0 & 473 \\
(f) & & \checkmark & \checkmark & \checkmark & \checkmark & \textbf{23.7} & \textbf{40.3} & \textbf{24.6} & 473 \\
\end{tabular}
\vspace{3pt}
\caption{
Ablation studies on the detection part on COCO test-dev.
We use a compressed Light-Head R-CNN with SNet146 as the baseline (BL), and gradually add small RPN (SRPN), small R-CNN (SRCN), Context Enhancement Module (CEM) and Spatial Attention Module (SAM) for ablation studies.
}
\label{table:ablation-study-architecture-modules}
\end{table}

We then insert Context Enhancement Module (CEM) after the backbone.
The output feature map of CEM is used for both RPN and RoI warping.
CEM achieves thorough improvements of 1.7 AP, 2.5 AP$_{50}$ and 1.8 AP$_{75}$ with negligible increase on FLOPs (Table~\ref{table:ablation-study-architecture-modules}(d)).
The combination of the multi-scale feature maps introduces semantic and context information of different levels, which improves the representation ability.

\vspace{-12pt}
\paragraph{Spatial Attention Module.}


\begin{figure}[!t]
\centering
\includegraphics[width=0.47\textwidth]{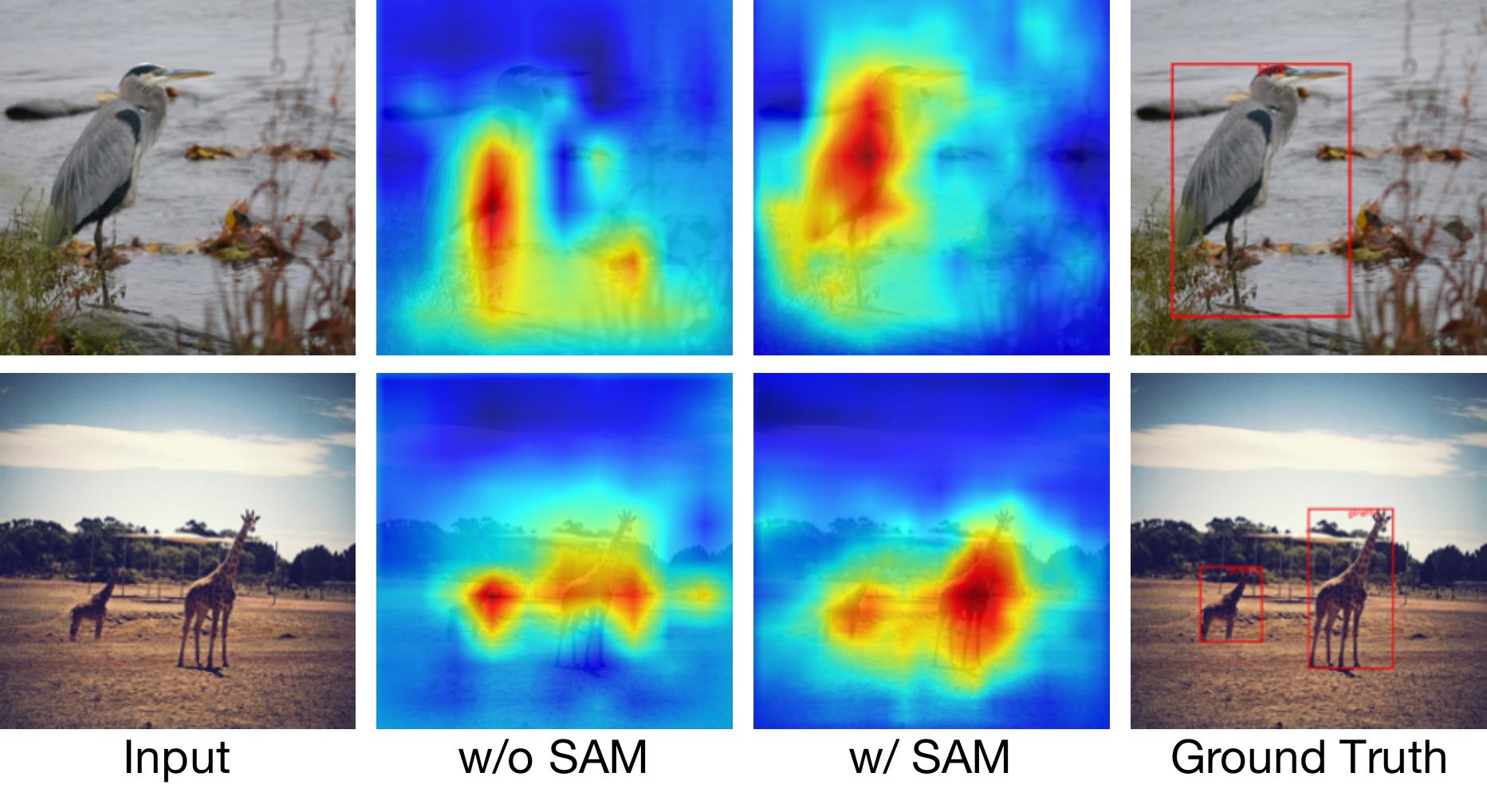}
\caption{Visualization of the feature map before RoI warping. Spatial Attention Module (SAM) enhances the features in the foreground regions and weakens those in the background regions.}
\label{figure:attention-feature-map}
\end{figure}

Adopting Spatial Attention Module (SAM) without CEM (Table~\ref{table:ablation-study-architecture-modules}(e)) improves AP by 1.4 with merely 5\% extra computational cost compared with Table~\ref{table:ablation-study-architecture-modules}(c).
Fig.~\ref{figure:attention-feature-map} visualizes the feature maps before RoI warping in Table~\ref{table:ablation-study-architecture-modules}(c) and Table~\ref{table:ablation-study-architecture-modules}(e).
It is clear that SAM effectively refines the feature distribution with foreground feature enhanced and background features weakened.

At last, we adopt both CEM and SAM to compose the complete ThunderNet (Table~\ref{table:ablation-study-architecture-modules}(f)).
This setting improves AP by 1.8, AP$_{50}$ by 2.7, and AP$_{75}$ by 2.1 over the baseline while reducing the computational cost by 34\%.
These results have demonstrated the effectiveness of our design.

\vspace{-5pt}
\subsubsection{Balance between Backbone and Detection Head}
\label{section:large-backbone-or-heavy-head}

We further explore the relationship between the backbone and the detection head.
Two models are used in the experiments: a \emph{large-backbone-small-head} (LBSH) model and a \emph{small-backbone-large-head} (SBLH) model.
The LBSH model is ThunderNet with SNet146.
The SBLH model uses SNet49 and a heavier head: $\alpha$ is 10, and a 2048-d \emph{fc} layer is used in R-CNN subnet.
As shown in Table~\ref{table:ablation-study-light-heavy-head}, the LBSH model outperforms the SBLH one by 3.4 AP even with less FLOPs.
It suggests that \emph{the large-backbone-small-head design is better than the small-backbone-large-head design for lightweight two-stage detectors}.
We conjecture that the capability of the backbone and the detection head should match.
In the small-backbone-large-head design, the features from the backbone are relatively weak, which makes the powerful detection head redundant.

\begin{table}[!t]
\scriptsize
\centering
\begin{tabular}{l|c|c|c|c|c}
Model & Backbone & RPN & Head & Total & AP \\ \hlineB{2.5}
large-backbone-small-head & 338 & 43 & 92 & 473 & 23.6 \\
small-backbone-large-head & 154 & 70 & 286 & 510 & 20.2 \\
\end{tabular}
\vspace{3pt}
\caption{
MFLOPs and AP of different detection head designs on COCO test-dev.
The large-backbone-small-head model outperforms the small-backbone-large-head model with less FLOPs.
}
\label{table:ablation-study-light-heavy-head}
\end{table}

\begin{table}[!t]
\setlength{\tabcolsep}{10pt}
\centering
\scriptsize
\begin{tabular}{l|c|c|c}
Model & ARM & CPU & GPU \\ \hlineB{2.5}
Thunder w/ SNet49 & 24.1 & 47.3 & 267 \\
Thunder w/ SNet146 & 13.8 & 32.3 & 248 \\
Thunder w/ SNet535 & 5.8 & 15.3 & 214 \\
\end{tabular}
\vspace{3pt}
\caption{
Inference speed in fps on Snapdragon 845 (ARM), Xeon E5-2682v4 (CPU) and GeForce 1080Ti (GPU).
}
\label{table:inference-speed}
\end{table}

\subsection{Inference Speed}
\label{section:inference-speed}

At last, we evaluate the inference speed of ThunderNet on Snapdragon 845 (ARM), Xeon E5-2682v4 (CPU) and GeForce 1080Ti (GPU).
On ARM and CPU, the inference is executed with a \emph{single thread}.
The batch normalization layers are merged with the preceding convolutions for faster inference speed.
The results are shown in Table~\ref{table:inference-speed}.
ThunderNet with SNet49 achieves real-time detection on both ARM and CPU at 24.1 and 47.3 fps, respectively.
To the best of our knowledge, this is the \emph{first} real-time detector and the \emph{fastest} single-thread speed on ARM platforms ever reported.
ThunderNet with SNet146 runs at 13.8 fps on ARM and runs in real-time on CPU at 32.3 fps.
All three models run at over 200 fps on GPU.
These results suggest that ThunderNet is highly efficient in real-world applications.


\section{Conclusion}

We investigate the effectiveness of two-stage detectors in real-time generic object detection and propose a lightweight two-stage detector named ThunderNet.
In the backbone part, we analyze the drawbacks in prior lightweight backbones and present a lightweight backbone designed for object detection.
In the detection part, we adopt an extremely efficient design in the detection head and RPN.
Context Enhancement Module and Spatial Attention Module are designed to improve the feature representation.
At last, we investigate the balance between the input resolution, the backbone, and the detection head.
ThunderNet achieves superior detection accuracy to prior one-stage detectors with significantly less computational cost.
To the best of our knowledge, ThunderNet achieves the first real-time detector and the fastest single-thread speed reported on ARM platforms.

\vspace{5pt}
\head{Acknowledgement.} This work is sponsored in part by the National Key R\&D Program of China (2018YFB2101100, 2017YFA0700800).


\begin{figure*}[!htb]
\centering
\subfigure{
\includegraphics[width=0.32\textwidth,height=0.24\textwidth]{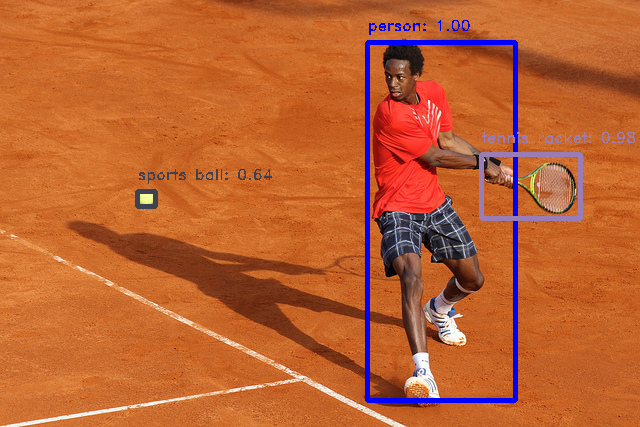}
}
\subfigure{
\includegraphics[width=0.32\textwidth,height=0.24\textwidth]{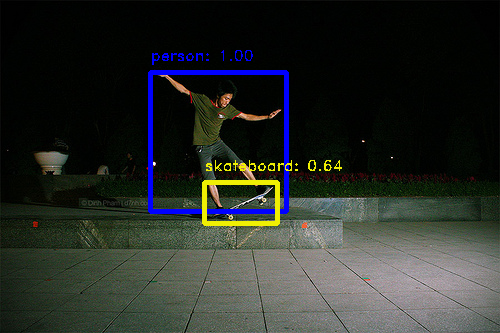}
}
\subfigure{
\includegraphics[width=0.32\textwidth,height=0.24\textwidth]{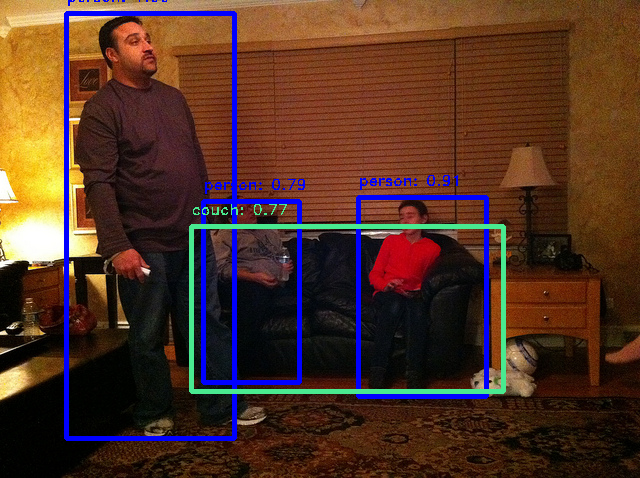}
}
\subfigure{
\includegraphics[width=0.32\textwidth,height=0.24\textwidth]{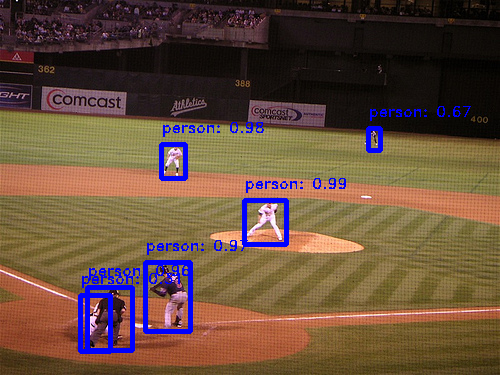}
}
\subfigure{
\includegraphics[width=0.32\textwidth,height=0.24\textwidth]{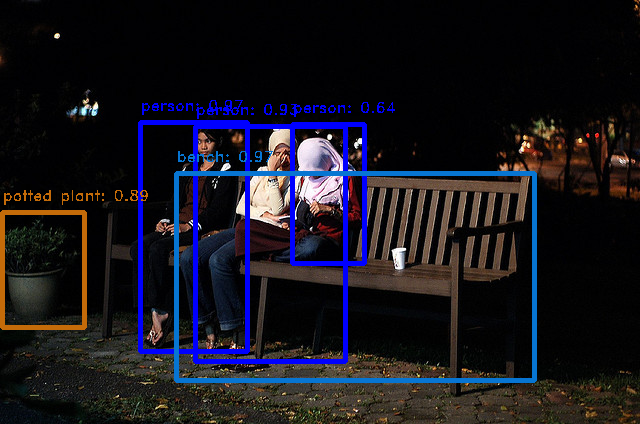}
}
\subfigure{
\includegraphics[width=0.32\textwidth,height=0.24\textwidth]{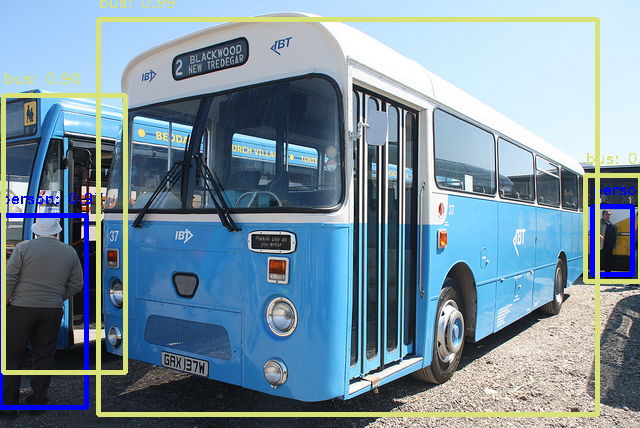}
}
\caption{Examples visualization of ThunderNet with SNet49.}
\end{figure*}

\begin{figure*}[!htb]
\centering
\subfigure{
\includegraphics[width=0.32\textwidth,height=0.24\textwidth]{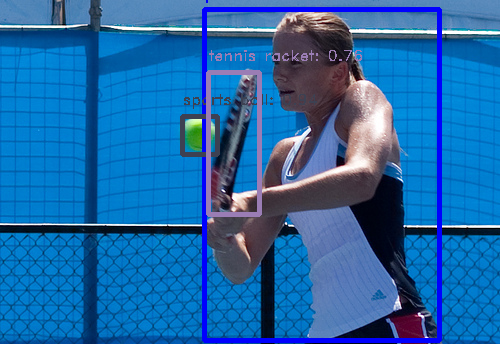}
}
\subfigure{
\includegraphics[width=0.32\textwidth,height=0.24\textwidth]{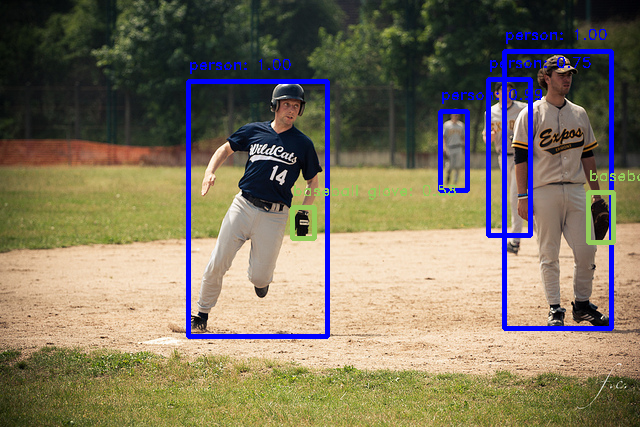}
}
\subfigure{
\includegraphics[width=0.32\textwidth,height=0.24\textwidth]{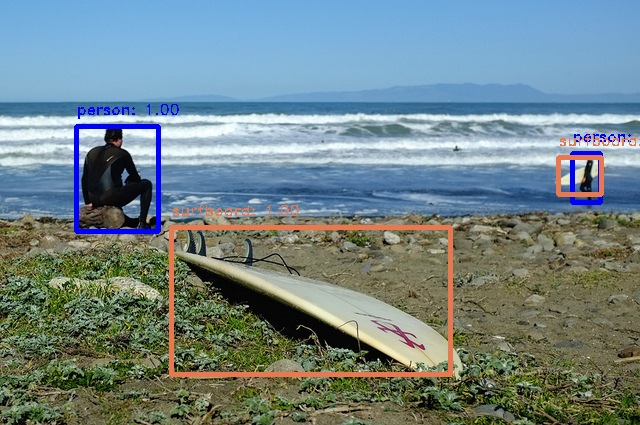}
}
\subfigure{
\includegraphics[width=0.32\textwidth,height=0.24\textwidth]{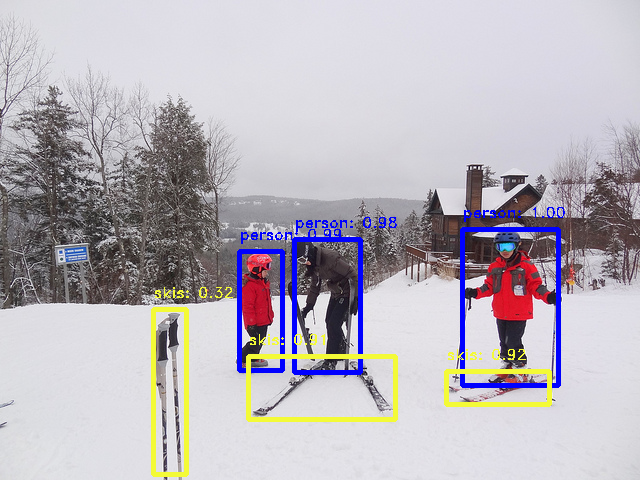}
}
\subfigure{
\includegraphics[width=0.32\textwidth,height=0.24\textwidth]{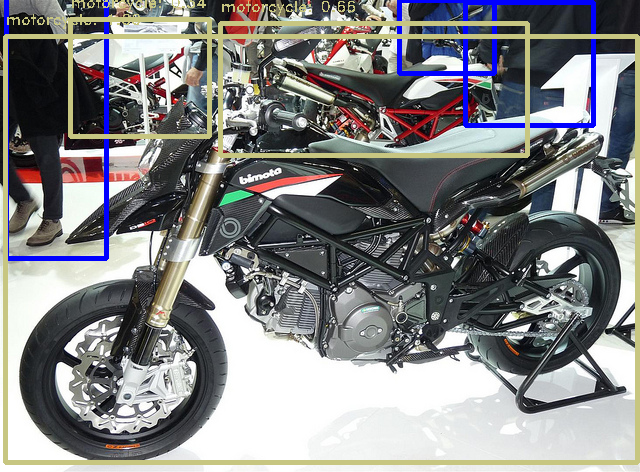}
}
\subfigure{
\includegraphics[width=0.32\textwidth,height=0.24\textwidth]{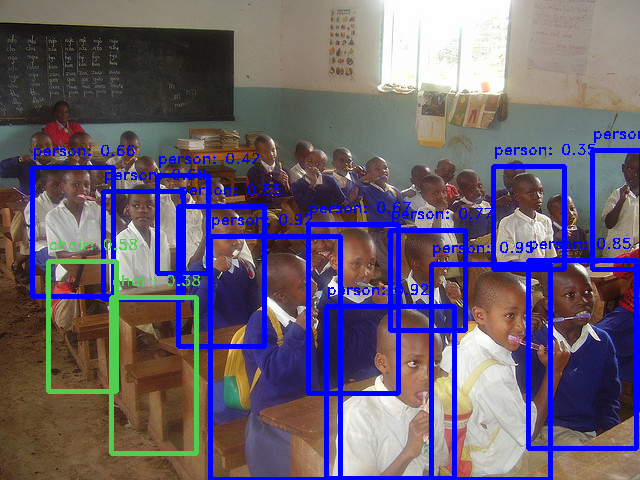}
}
\caption{Examples visualization of ThunderNet with SNet146.}
\end{figure*}

\begin{figure*}[!htb]
\centering
\subfigure{
\includegraphics[width=0.45\textwidth,height=0.3\textwidth]{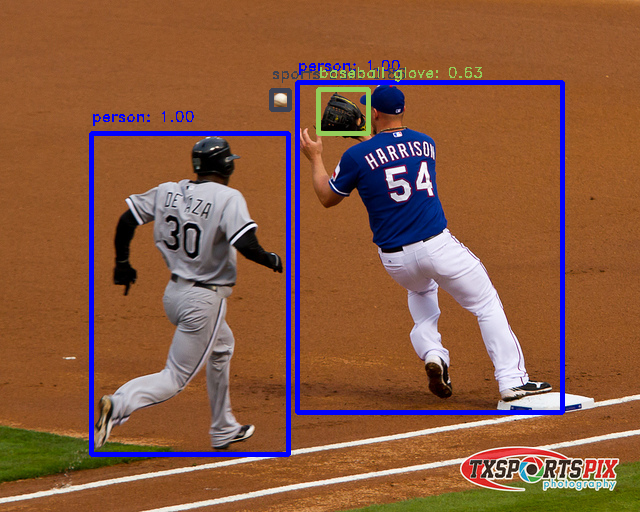}
}
\hspace{10pt}
\subfigure{
\includegraphics[width=0.45\textwidth,height=0.3\textwidth]{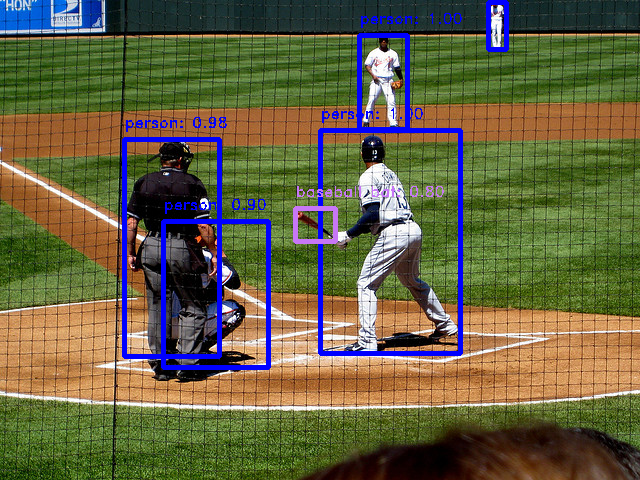}
}
\subfigure{
\includegraphics[width=0.45\textwidth,height=0.3\textwidth]{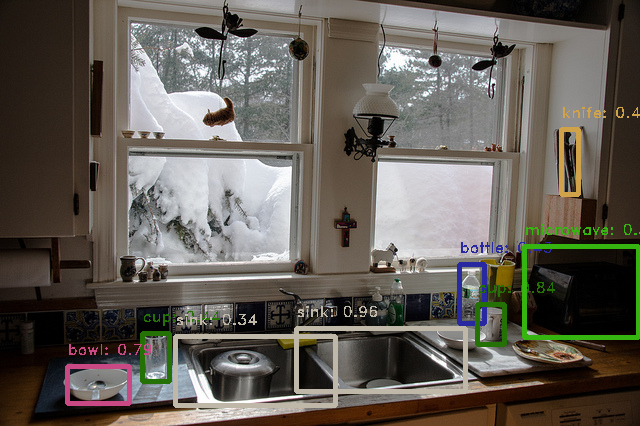}
}
\hspace{10pt}
\subfigure{
\includegraphics[width=0.45\textwidth,height=0.3\textwidth]{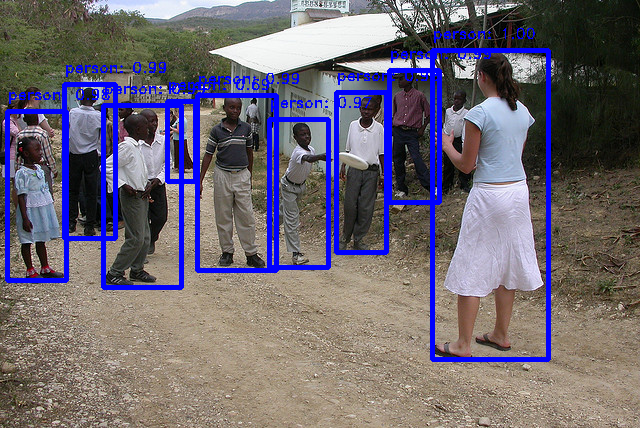}
}
\subfigure{
\includegraphics[width=0.45\textwidth,height=0.3\textwidth]{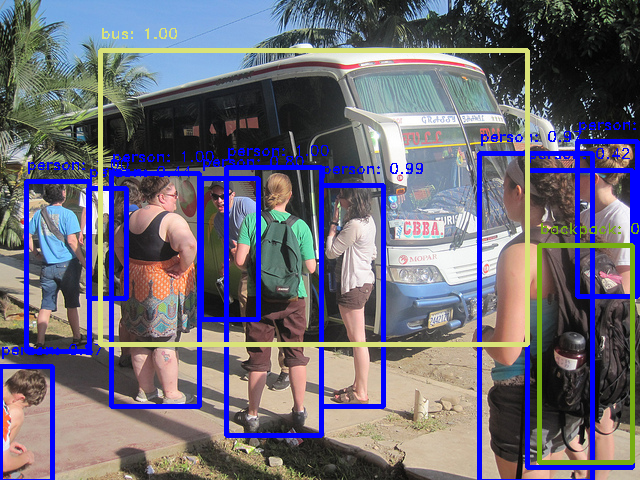}
}
\hspace{10pt}
\subfigure{
\includegraphics[width=0.45\textwidth,height=0.3\textwidth]{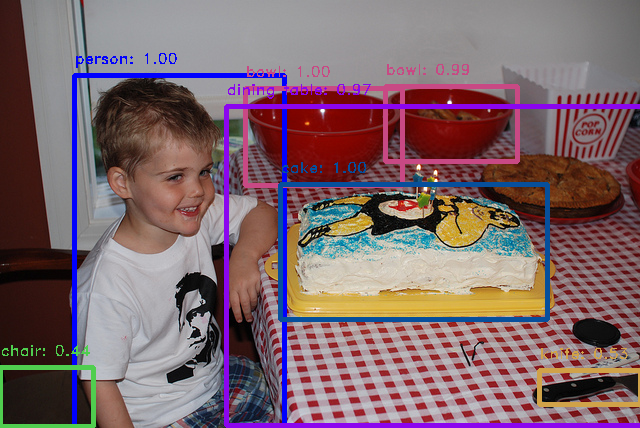}
}
\subfigure{
\includegraphics[width=0.45\textwidth,height=0.3\textwidth]{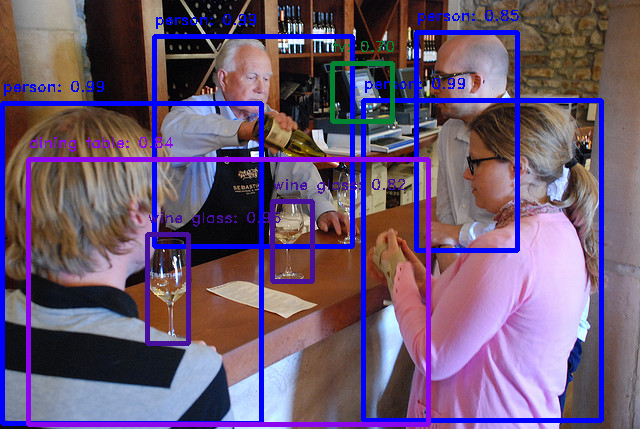}
}
\hspace{10pt}
\subfigure{
\includegraphics[width=0.45\textwidth,height=0.3\textwidth]{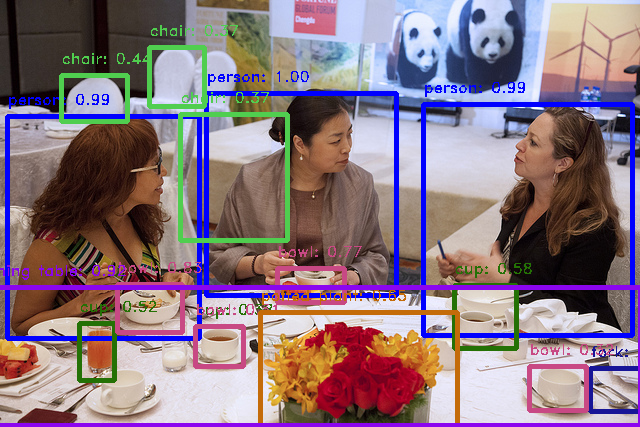}
}
\caption{Examples visualization of ThunderNet with SNet535.}
\end{figure*}

{\small
\bibliographystyle{ieee_fullname}
\bibliography{thundernet}
}

\end{document}